# Dense Sample Deep Learning

*Authors: Stephen Josè Hanson , Vivek Yadav,  & Catherine Hanson*
Affiliation: Rutgers University, RUBIC, Psychology Department, CMBN

*Dense Sample Deep Learning*

Abstract

Deep Learning, a variant of the neural network algorithms originally proposed in the 1980s, has made surprising progress in Artificial Intelligence (AI), ranging from language translation, protein folding (Jumper et al, 2021), autonomous cars, and more recently human-like language models (CHATbots), all that seemed intractable until very recently. Despite the growing use of Deep Learning (DL) networks, little is actually understood about the learning mechanisms and representations that makes these networks effective across such a diverse range of applications. Part of the answer must be the huge scale of the architecture and of course the large scale of the data, since not much has changed since 1987. But the nature of deep learned representations remain largely unknown. Unfortunately training sets with millions or billions of tokens have unknown combinatorics and Networks with millions/billions of hidden units can't easily be visualized and their mechanisms can't be easily revealed. In this paper, we explore these questions with a large (1.24M weights; VGG) DL in a novel high density sample task (5 unique tokens with 500+ exemplars per token) which allows us to more carefully follow the emergence of category structure and feature construction. We use various visualization methods for following the emergence of the classification and the development of the coupling of feature detectors and structures that provide a type of graphical bootstrapping, From these results we harvest some basic observations of the learning dynamics of DL and propose a new theory of complex feature construction based on our results.

*Dense Sample Deep Learning*

*Dense Sample Deep Learning*

*Introduction: human categorization learning*

Perhaps the most common task that Deep Learning models have been successful with is classification and most frequently using image data.  This kind of task in cognitive science might be termed categorization or concept learning (Shepard, Hoveland & Jenkins, 1961; Hanson & Gluck, 1990; Bruner, Goodnow and Austin, 1959) or more fundamentally *identification*   (Luce, Bush, Galanter, 1963).   Although there are similarities to human learning, there are some important differences with typical DL classification.   One key difference involves the distribution of exemplars (samples) and the level at which they are sampled.  Specifically, in cognitive science, the level of a category (Roche et al, 1976), is based on the hierarchical nature of category and a preferred level of reference (e.g. Basic level and will vary across individual expertise). More critically in supervised learned in DL architectures the nature of the representation is dependent on the similarity function learned that maximally separates within category members from between category members.  Because the DL like any neural network is a general function approximator ( Hanson & Burr, 1990, Carroll et al, 1989; Honik et al, 1990), it is difficult to tell whether the similarity functions learned for the mapping are arbitrary or are consistent with human bias (Hanson et al, 2019)   Supervised learning using DL usually involves labeled data  that may only sparsely represent any "concept" or category  given the arbitrary category label.  These type of classification tasks are often used to  establish benchmarks with previously developed shared image databases in order to compare algorithm variations, for example,  with CIFAR (60k images, where there are 10 or 100 categories making the exemplar sample space sparse) and similar kind of diversity with ImageNet (14M images; based on G. Miller's WordNet; although often used with a reduced sample of 1.28M images).   For CIFAR the category structure  in both labled datasets (10,100) is diverse and sometimes appears at the basic level ("fish", "man" etc..) and sometimes as some non-familiar subordinate level ( "aquarium fish"--as opposed to say  "goldfish") and also cases containing mixed levels of reference.   The other common category learning dataset is the ILSVRC subset of ImageNet, which  contains around 1.28 million training images representing 1,000 categories. These categories also tend to be at both the basic and subordinate levels as well,  with a focus on fine-grained classification but across multiple levels of reference.  Hence,  in these typical benchmark data sets, there doesn't appear to be common and consistent level of reference relative to human lexical knowledge or usage.  Which means the concept space can be very sparsely covered, although there will many exemplars clustered with large gaps throughout the feature hyper-cube where the exemplars distribute.  But of course, having benchmark datasets that are  consistent with human bias was not a primary goal, nor necessary for comparative tests of architectures or learning algorithms.

We introduce here, a more ecologically valid human learning task that involves  a smaller set of categories but with a dense set of exemplars per category.  Similar to the nearest kin and friends and famous individuals (actors, politicians etc.) that a single individual might have. We term this the dense sample category task (DSC).

One such set is the *Yale Face Data*, where each face has no less then 500 exemplars per face (based on 5760 images of 10 individuals, each under nine poses and 64 different lighting conditions-576 cases). This makes the concept space very dense in a feature hyper-cube, since it involves a singular category, a single identity. Of course, there are categories with 1000s of exemplars in either CIFAR or ILRCVC but at a much coarser level and never with this kind of density *per concept*.

We randomly chose five faces (although human subjects can recognize more than 1000 of faces accurately) and as pointed out this tactic might represent immediate family members in that the individual will have a constant and extensive exposure. Thus the goals of this study, are to (1) approximate the sample space of a typical human individual with a similar density and learning exposure. (2) To be able to compare and contrast the feature construction in the hidden layers early and late in learning as well as early and late topographically in the network. (3) To analyze the dynamics of learning and to characterize learning phases when features first "crystallize", i.e., when category formation first appears. (4) To determine whether learning is unitary or multi-factorial.

**Methods and Materials:**

We trained a full DL model (2.4M weights; see table 2) and focus on the layer to layer interactions in order to characterize the internal learning dynamics. Faces were randomly selected from the 10 set of faces. These shown in figure 8 are typical of the variation of the entire set and show a strong similarity across individuals (women were underrepresented in the set), leading to 1 female and 4 males in our selected set.

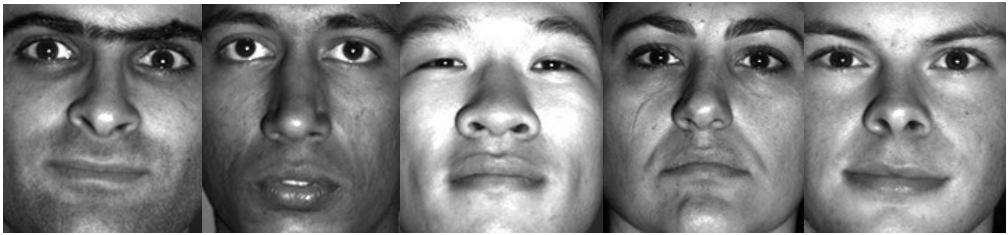

Figure 8. The face images of the five individuals in the database after a close crop removing background features.

In order to identify the network architecture for this data we performed several empirical experiments on a VGG style network (starting with a VGG-11) and was able to shrink to a network with five convolutional layers with kernel size three and three filters per layer. Decreasing the size of the network any further made the learning process very unstable and proved to be difficult for getting consistent results while experimenting. We also used the smallest network size possible so as to better understand the functionality of each of the layers and to reduce unnecessary parameters given the small and condensed size of our classes. We also used a fixed filter size for all the convolutional layers so the differences between the layers could

```
----------------------------------------------------------------
        Layer (type)               Output Shape         Param #
================================================================
            Conv2d-1          [-1, 3, 128, 128]              84
              ReLU-2          [-1, 3, 128, 128]               0
            Conv2d-3          [-1, 3, 128, 128]              84
              ReLU-4          [-1, 3, 128, 128]               0
            Conv2d-5          [-1, 3, 128, 128]              84
              ReLU-6          [-1, 3, 128, 128]               0
            Conv2d-7          [-1, 3, 128, 128]              84
              ReLU-8          [-1, 3, 128, 128]               0
            Conv2d-9          [-1, 3, 128, 128]              84
             ReLU-10          [-1, 3, 128, 128]               0
 AdaptiveAvgPool2d-11           [-1, 3, 20, 20]               0
           Linear-12                  [-1, 1024]       1,229,824
             ReLU-13                  [-1, 1024]               0
          Dropout-14                  [-1, 1024]               0
           Linear-15                  [-1, 1024]       1,049,600
             ReLU-16                  [-1, 1024]               0
          Dropout-17                  [-1, 1024]               0
           Linear-18                     [-1, 5]           5,125
================================================================
Total params: 2,284,969
Trainable params: 2,284,969
Non-trainable params: 0
----------------------------------------------------------------
Input size (MB): 0.19
Forward/backward pass size (MB): 3.81
Params size (MB): 8.72
Estimated Total Size (MB): 12.71
----------------------------------------------------------------
```

be better analyzed and understood. This left us with images of image size 128x128 at the end of the convolutional stages and with further focused pooling kept the size of the model manageable in the fully connected upper layers of the DL.

*Factoring learning dynamics.*

Here we will describe in more detail this mechanism, using this specific archive data set (the Yale Face data). Consider, for example, the two dimensional PCA representation of the exemplar distribution during learning (Figure 9 and 10).

The first figure (9) shows the beginning of learning and the lack of separation of the five face categories. The error by chance is 22% (1/5). The second figure (10) on the right shows learning at asymptote, after 51 epochs, showing the five categories completely separated (acc=97%) in the 2 dim PCA space.

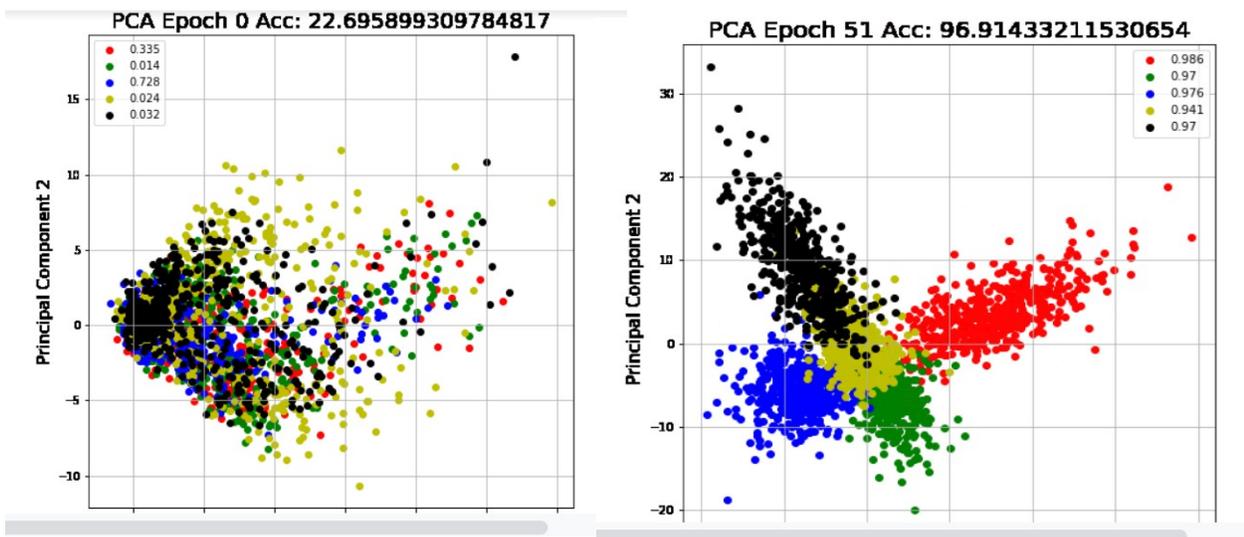

Next, we analyzed the resultant learning curve shown in figure 11, where we have fit with a sum of two logistic curves. In this case, a single logistic curve fits with around 94% of the overall variance but with systematic errors on the phase transition when the first face identification occurs, while the second logistic (albeit with 2 more parameters), boosts the fit to 99% of the variance, with a visually unbiased fit over the whole learning function.

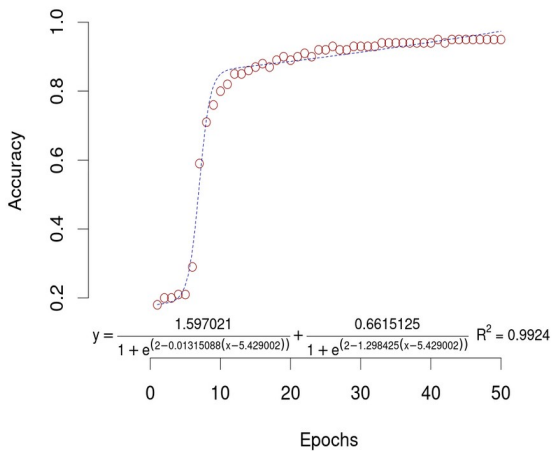
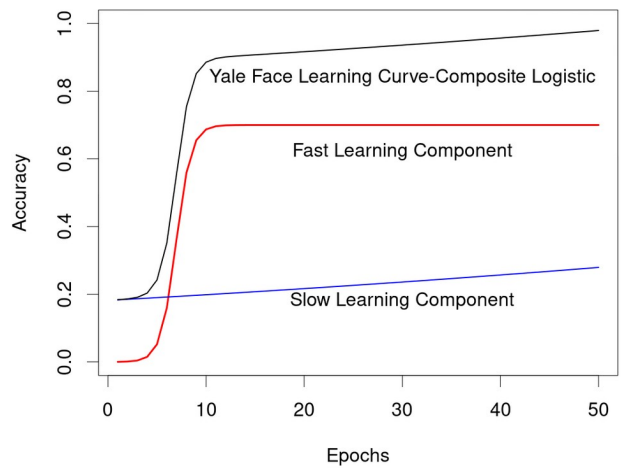

*Figure 11: Cumulative Logistic fit to Yale Face Data, DL Learning curve, note the two components logistic curves below*

*Figure 12: Factoring the components underlying Yale Face data DL learning curve*

We will argue this factorization of learning dynamics is the key to understanding dynamics in deep learning. Specifically, we propose that multiple learning processes tend to be initiated as a series of hyperbolic learning processes. As in a *wavelet* decomposition or really, any type of spectral decomposition, we hypothesize Deep Learning dynamics can also be decomposed in a series of hyperbolic functions, some very fast, some slower and others near the floor—effectively background processes—but ones that are no less critical to the final representation. We will term this factoring the logistic learning decomposition (LLD):

(13) $$\sum_{i}^{n} C_i a_i / (1 + e^{-b_i})$$

*Dense Sample Deep Learning*

The number of terms in this series will be a function of the number of categories, the complexity of the decision surface (non-linearity, convexity, connectedness etc..) and data complexity. In terms of this small number of categories, 2 factors are not unreasonable, however, we would expect in CIFAR there much be 100s of such hyperbolic processes—some of which must be highly correlated (see below). Fast hyperbolic processes, we will see next are based on the highest variance extraction which will usually be based on the first significant class separation, in this case the RED class- "face 1" exemplars, but could based on the entire linear separability of the task—say the Fisher's IRIS data-- two decision surfaces might be extracted close in time, producing 2 large variance spikes in the learning dynamics (the third one which is more nonlinear taking a slower process)—with at the same time a fairly coarse structure. More moderate hyperbolic learning processes in DL can therefore independently adjust the decision surface gradually improving the overall accommodation of other exemplars, while slower hyperbolic processes can discover more complex, higher fidelity feature detectors and structure in the data producing the smoothest and highest probable fit to the true decision surface. It is also important in order to understand that the learning dynamics will tend to be a *sequential* extraction (similar to PCA, but not necessarily with orthogonal components—in fact unlikely), making the overall DL approximation a series of conditionally independent feature detectors/filters (partly due to the spectral separation resulting from depth of each layer) that can be incrementally added to a larger hierarchy or "scaffold" that may have already developed.

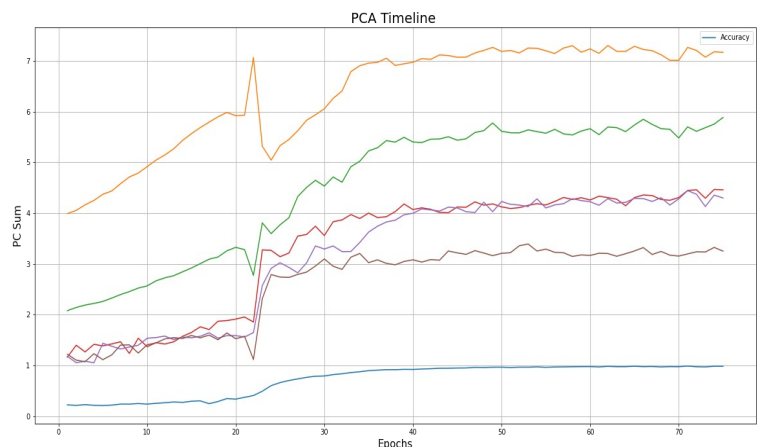

Figure 13: Principal components analysis over last 1024 hidden units prior to classification ---blue line is total error over epochs.

*Looking further inside the black-box*

In order to further back-fill this type of mechanism we explore, next the representation of the five face categories at the change-point where learning abruptly ends. a latent phase and increases hyperbolically, for reasons we have outlined earlier we consider this to be the "accumulation" phase of the learning, as we will see there is a sequence of learning structures supporting the final classification. Consider in figure 13 where we model the dynamics of the last 1024 hidden units prior classification. We performed a PCA over all classes and exemplars on the activation of hidden units at each epoch. As learning proceeds we see the composite score of the largest PC (we extract 5 PCs that represents near 99% of the entire variance of the set) peak near epoch 22, shows an emergence of an entire class of faces (Figure 14: Face1-RED exemplars), with the others emerging near epoch 26 and finally by epoch 30. At the point of transition there is a large spike in the hidden layer accommodating the convolutional layers that have created feature detectors that through the first 22 epochs were not sufficiently tuned to separate a critical mass of the cases, in effect the category representation of face 1.   We show for face 1 and face 5, the pixel-wise variance across all exemplar faces per category, and further show likely hypotheses about feature detectors as they are forming.  Here in Figure 15, are nose, eye brow and cheek regions, each the basis for a larger unique set of detectors that will eventually emerge.   Note in Figure 16 the other extracted PCs, show a perturbation at the point of phase transition, but with  smaller peaks and depressions prior to a steady rise in variance accumulation.   The bottom blue curve shows the accuracy as the DL approaches asymptotic classification and the change point near epoch 22-28.

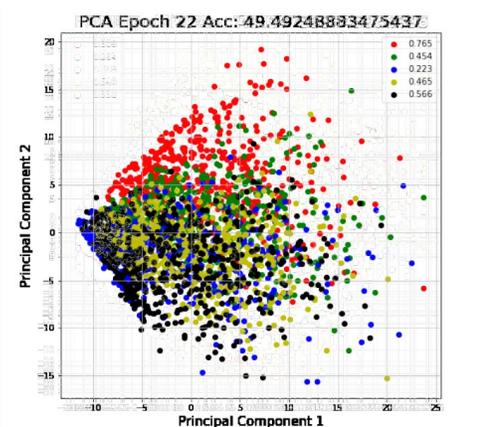

*Figure 14: First appearance of one face category (red) around epoch 22.*

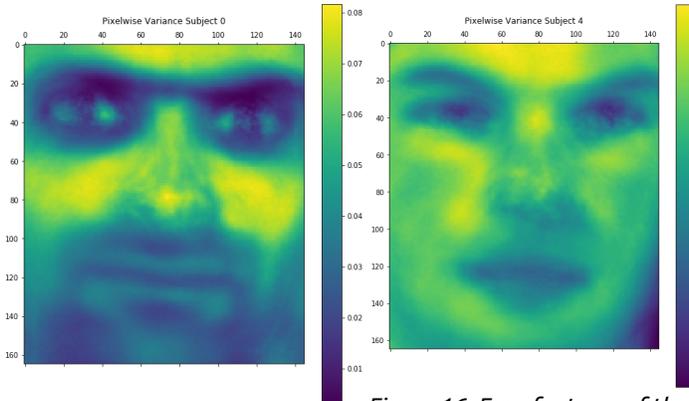

*Figure 15: Face features at  for first emerging face category*

*Figure 16: Face features of the fifth emerging face category.*

It is clear that the hidden units form different paths as they cluster together in various combinations within each PC.  This is a type of competition which produces sets of PC that are mostly uncorrelated.   These feature combinations (PCs) are initially hypotheses about the the classification of the categories, in the prototype cases for each of  the 5 faces.    The question of how independent the hidden layers have become are not obvious from the components alone. We, therefore also calculated the correlation pairwise of all convolutional layers over epochs shown in Figure 17. Similar to the PCs over time, there is a phase transition around epoch 22-28 and after that point there is a clear orthogonality of the hidden layers throughout the network although with the spatially closest hidden units with higher correlation and those further away with lower correlations. This gradient was nearly linear over the 5 layers, with correlations of spatially close layers are slowly drifting down, while those that are spatially further away are decreasing more quickly and steadily over epochs. It is clear that a specific weight correlation structure per layer is at the basis of these hidden unit functions and the dynamics of the DL learning. Nonetheless, it is also clear that the correlation structure of the DL is complex at the layer/weight level.

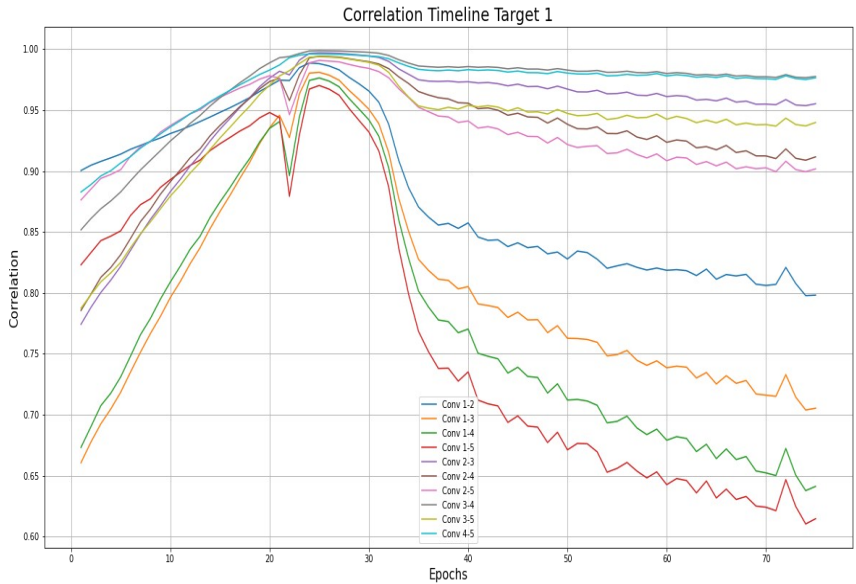

Figure 17: Correlation over epochs of successive pairs of layers over DL model

*Dense Sample Deep Learning*

**Discussion**

One recent theoretical (Martin & Mahoney, 2021) approach to understanding the dynamics of DL learning is to frame the weight dynamics as a disordered system using statistical mechanics. In this theory, the over parametization paradox of DL is characterized as a natural regularization of parameters in DL networks through competitive implicit process (thus no explicit regularization; c.f. Moody, 1991). Regularization of models in the standard case (Tikonov regularization; NN weight decay) involves adding a penalizer to the estimator so that parameter dynamics are damped and the parameter efficiency is optimized (Hanson & Pratt, 1989). This new theory of regularization was constructed on 100+ DL learned architectures by extracting the eigenvalue distribution from the correlation of weight matrices per layer. What resulted was a theory of DL learning that shows typically, 5 distinct phases of learning. Initially there is a RANDOM eigenvalue distribution that slowly evolves to have developed a new theory of regularization (heavy-tailed regularization as in contrast to classical regularization-Tikonov) that appears to apply to deep-learning architectures. The theory predicts 5 phases of learning in terms of the Eigenvalue spectral distribution (ESD) including RANDOM, BULK, BULK-bleedout, BULK with spikes, and final rank collapse. The term BULK refers to the eigenvalue distribution that results with weak covariance soon after random ESD distribution appears (which would imply no learning). The "bleedout" and "spikes" are the first signs of a strong covariance structure emerging, which then rapidly grows to a highly connected predictive structure prior to the rank collapse. We will return to this theory as we discuss a new proposal for DL feature construction. But first lets consider what we can say so far about deep-learning dynamics.

*Harvesting some principles of deep learning dynamics*

During learning and visualization we observed a number of regularities which we will summarize next in the context of at least two kinds of learning that we have observed in this example, what we will define as "fast feature competition" and "slow feature curation".

(1) **more layers create a "buffer" from aggressive competitive learning, common in single layer networks.** The layers in a DL appear locally correlated suggesting that competitive learning which is common in perceptron and backpropagation single hidden layer networks is isolated per layer thus slowing down destructive/constructive learning throughout later layers.

(2) **layers can enable a "latent learning", or induction period, to more comprehensively model data complexity** with smaller less destructive (near zero) gradients will preserve promising feature analyzers in later layers, and in effect "curate" them with more consistent samples with more consistent feature sets that middle layers will filter.

(3) **layers decouple long range effects and effectively create a conditionally independent network of layers.** This effect allows for more local updates per layer conditioned on the spatially nearest layers, thus again increasing the locality and ultimate fidelity of the feature analyzers and protecting them from descrutive competiton.

(4) **As slow logistic processes increase in accuracy, a tipping point is reached with an explosive growth of accuracy reflecting the network of feature analyzers that are nearly similar to what they will be asymptotically.** This is the logistic or exponential-hyperbolic form has a slow rise period (induction period) prior to explosive growth with slower refinement processes (Hanson et al, 2019; Hanson et al, submitted).

(5) **There are multiple learning processes initiated in parallel but at different rates depending on covariance/complexity in the data.** These learning processes will accumulate to an overall learning curve which can be factored into N multi-logistic learning functions with different rates (as shown earlier). The faster processes during classification, are extracting lower dimensional (separable features) structure while slower processes are back-filling structure in from covariance (integral features), similar to a nonlinear factor analysis (Shepard et al 1970). These slower logistic processes might be termed, latent "correct" responses, which emerge over a longer resolution period and at the same time with a much faster, hyperbolic rise in learning. Clearly, DL is multi-factorial.

(6) **We would expect the deeper the network the more quickly destructive competition and slow "curation" of feature structures are trading off.** Consequently, lower layers (closest to input) may experience more rapid change in accommodating the higher layer's incremental improvement of feature detectors that tend to be more consistent with error feedback outcome per sample/batch. This potentially creates a long distance communication channel between lower and upper layers. Nonetheless, there must be diminishing returns on this strategy and the effective depth of any DL learner.

**A new theory of feature creation in DL: Auto-catalytic feature sets.** As we have seen the logistic (hyperbolic) function can produce explosive growth to an asymptote with a complex construction of novel feature analyzers. We propose a theory of feature creation in DL networks that is based on an analogy to a chemical reaction. Recall that catalysis is typically described as a chemical reaction that normally proceeds at a fixed rate, depending on the chemical constituents present. A catalyst will multiply the basic rate of that reaction by many fold, thus increasing the rate of the reaction without itself being consumed in the reaction. The resultant product is drastically increased in the presence of the catalyst. The graph for these equations is a sigmoid curve (specifically a logistic function), which is typical for auto-catalytic reactions: these chemical reactions proceed slowly at the start (the induction period) because there is little catalyst present, the rate of reaction increases progressively as the reaction proceeds as the amount of catalyst increases and then it again slows down as the reactant concentration decreases.

Autocatalytic sets (c.f. Kaufmann, 1992), as may be apparent from their name are that they literally catalyze themselves, in effect, auto-catalytic sets produce their own catalyst during their reaction, providing cross catalysis for connected variables (in the case of a dynamic system like Lotka Volterra, it inhibits and therefore symmetrically dis-inhibits the growth of the other). The second property from Kauffman's original theory, is that an auto-catalytic set will appear as one "giant connected component" in the chemical reaction.   In effect the autocatalytic set is a kind of "clique" in a graph or a circuit  which provides for the sustainability and **hyperbolic** growth of structure as it develops with  more and more elaborate structure, what we referred to earlier as "curated" features.

One of the continuing mysteries of DL feature construction is how "raw", basic features input to a convolutional layer becomes some sort of high fidelity representation of a "face", "cat", or  "car".  Are there universal structures that  evolve for the recognition of types and tokens?   Auto-catalytic sets may also account for how  they can evolve at all from such a simple pixel level input,  For example, recent work in DL visualization has revealed complex features, that are not merely obvious bits and pieces of original feature structure from the stimulus set, but rather, appear to be based on inferred  qualities of symmetry, relational structure, texture, color, pattern and  central features. They appear to be novel inventions of the DL (see figure 20; nose, eyes in case of the "dog" category; Olah et al 2017; 2020).

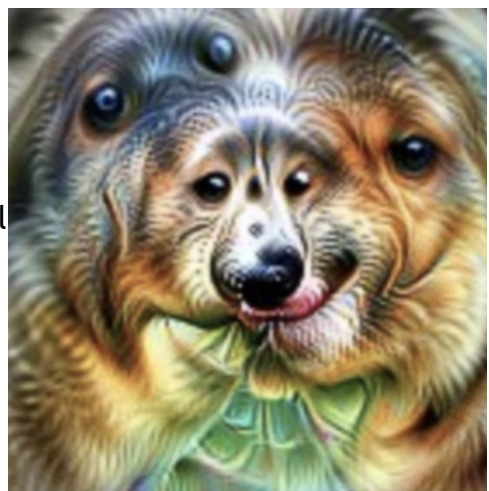

*Figure 20:  Visualization of large scale DL Olah 2020.   "Dog" category.*

In comparison this type of feature represetation is also still a central  mystery in visual pathways in the brain which in standard neuroscience  textbooks typically show pathways with simple features (Hubel & Wiesel, 1954),  edges and lines, to patterns and textures and more complex checkerboards and then-- *dot.. dot.. dot*-- to objects, faces, and complex types, but no *specific intermediate structures* have been observed specified theoretically or identified that might bootstrap or provide constructive routes to more complex forms, its only just a plausible hypothesis. Worse, It has been conventional wisdom that hierarchical structures *could* emerge that *could* in principle build and be also diagnostic of any type or token that might be in the world (e.g., Biederman, 1987). But these proposals have generally not been productive and are rife with many types of inductive, constructive, and logical problems (Herzog et al , 2014; Dickenson et al,.1997; Edelman,1997). The dorsal visual pathway of the brain that may be primarily focused on "what" information at some system level, but is still poorly understood on *what* the "what" pathway actually does. It seems ironic, that instead of the usual trope of technology rushing ahead of human common usage, we have created artificial systems based on the human brain as complex as the human brain, creating yet another mystery to investigate.

*Dense Sample Deep Learning*

*Dense Sample Deep Learning*

**Conflict of interest:  None, All authors**